\documentclass{IOS-Book-Article}

\usepackage{mathptmx}
\usepackage{url}
\usepackage{graphicx}

\def\hb{\hbox to 10.7 cm{}}

\begin{document}

\pagestyle{headings}
\def\thepage{}

\begin{frontmatter}              

\title{Assessing the Lockdown Effects on Air Quality during COVID-19 Era}

\markboth{}{May 2021\hb}



\author[0]{\fnms{Ioannis Kavouras}}
\author[0]{\fnms{Eftychios Protopapadakis}}
\author[0]{\fnms{Maria Kaselimi}}
\author[0]{\fnms{Emmanuel Sardis}}
and
\author[0]{\fnms{Nikolaos Doulamis}}

\address[0]{National Technical University of Athens, Zografou Campus 9, Iroon Polytechniou str, Athens, 15780, Greece, 
ikavouras@mail.ntua.gr, eftprot@mail.ntua.gr, mkaselimi@mail.ntua.gr, ndoulam@cs.ntua.gr}

\begin{abstract}

In this work we investigate the short-term variations in air quality emissions, attributed to the prevention measures, applied in different cities, 
to mitigate the COVID-19 spread. In particular, we emphasize on the concentration effects regarding specific pollutant gases, such as carbon monoxide 
($CO$), ozone ($O_3$), nitrogen dioxide ($NO_2$) and sulphur dioxide ($SO_2$). The assessment of the impact of lockdown on air quality focused on four 
European Cities (Athens, Gladsaxe, Lodz and Rome). Available data on pollutant factors were obtained using global satellite observations. The level of 
the employed prevention measures is employed using the Oxford COVID-19 Government Response Tracker. The second part of the analysis employed a variety 
of machine learning tools, utilized for estimating the concentration of each pollutant, two days ahead. The results showed that a weak to moderate 
correlation exists between the corresponding measures and the pollutant factors and that it is possible to create models which can predict the behaviour 
of the pollutant gases under daily human activities.
\end{abstract}

\begin{keyword}
environmental quality\sep air pollution\sep air quality levels\sep neural networks\sep COVID-19\sep machine learning\sep dynamic time warping
\end{keyword}
\end{frontmatter}
\markboth{May 2021\hb}{May 2021\hb}
\thispagestyle{empty}
\pagestyle{empty}

\section{Introduction}

By the end of January 2020, authorities, worldwide, introduced a series of measures, in order to reduce the transmissibility of the COVID-19 virus epidemic \cite{Tosepu_2020}. These measures include international travel controls (e.g. travel prohibition between countries), restrictions in internal movements (e.g. restrictions in transportation between municipalities), public event 
cancellations (e.g. closure of live concerts), restriction gatherings (e.g. limiting the amount of people over an area), closure of public transportation, closure of schools, stay home requirements (e.g. move outside the house only for work, emergency, etc) and closure of workplaces (e.g. closure of industries / commercial sector) \cite{SITE_FOR_COVID_MEASURES_INTERNATIONALLY}.

The response measures appeared to have an impact on the Air Quality Levels (AQL) in cities. Indicative examples involve countries such 
as China \cite{SHARMA2020138878}, India \cite{GAUTAM202015}, Brazil \cite{NAKADA2020139087} and Spain \cite{TOBIAS2020138540}. However, 
there are no clear indication regarding the correlation between each major pollutant factor, i.e. Carbon Monoxide ($CO$), Ozone ($O_3$), 
Nitrogen Dioxide ($NO_2$) and Sulphur Dioxide ($SO_2$), and the applied measures. The general trends indicate that applied measures improved the AQL. 
There  are various environmental models, based on simulations \cite{ElectricVehicles2015, FERRERO2016450}, indicating the causality 
between specific actions and AQL. Nevertheless, the models have not been sufficiently tested in real life. Thus, the lockdown, worldwide, 
created a unique opportunity for researching the significance of traffic reduction in AQL changes.

In this work we will investigate the correlation between the applied lockdown measures and the four major
pollutant factors ($CO$, $O_3$, $NO_2$ and $SO_2$) for a total of four European Cities - Athens (Greece), 
Gladsaxe (Denmark), Lodz (Poland) and Rome (Italy) - during COVID-19 period. In particular, we focus on 
the identification of each measure and what scale affects each of the pollutant factors. This will be 
achieved by a variety of statistical correlation methods, where we will test the relation between each 
measure, at a time, and the density of each pollutant factor, for a yearly period divided by two days 
average values. The pollutant factor densities will be calculated using the Sentinel-5 TROPOMI satellite
images of a two day period, clipped in a squared area of radius $0.25^o$ around the center of each city.

Also, a variety of machine learning methods will be trained using the most correlated COVID-19 measures
and the densities of each one of the most dangerous pollutant gases ($CO$, $O_3$, $NO_2$, $SO_2$). A
different model will be created, for each of the pollutant factors. In total, four models per machine 
learning methods will be trained. As input we will use a two days period COVID-19 measures and each pollutant's
mean density. The output will be each pollutant's mean density, two days ahead.

\section{Related Work}
The multiple and synchronized applied measures (lockdown) for the minimization of COVID-19 transmissibility, 
provided the opportunity for an in depth research of the impact of industrial activities closure and traffic
volume reduction, in cities, and AQL. During and after the lockdown, in China, researches proposed a reductive
trend in $CO$, $O_3$, $NO_2$, $SO_2$, $PM_{10}$ and $PM_{2.5}$ densities 
\cite{ChinaAirPollution, HeChinaPollution, ChanchanGao_COVID-19}. Indicative example, consists the Hubei region,
in which during the lockdown period in January and February, 2020, the Angstrom Exponent (AE) and the 
Aerosol Optical Depth (AOD) increased and decreased respectively by 29.4\% and 39.2\% (only in Wuhan they increased
and decreased by 45.3\% and 31.0\%) \cite{SHEN2021142227}.

Another, significance example, includes India, in which has been observed reduction, during and after the lockdown,
in critical pollutant factors \cite{singh_diurnal_2020_IndiaPollution, shehzad_impact_2020, mahato_effect_2020_IndiaPollution}.
A comparison in air pollution changes between India and China, before and after the lockdown, described in the work of
\cite{agarwal_comparative_2020_ChineseIndiaPollution}. Real-time measured densities of $PM_{2.5}$ and $NO_{2}$ are used for a 
total of 12 megacities in India and China (6 megacities from each country). An Air Quality Index (AQI) is calculated, for each
density, before, during and after the lockdown, which resulted to a to a reduction of 45.25\%($AQI_{PM_{2.5}}$) and 
64.65\%($AQI_{NO_{2}}$) in China and 37.42\%($AQI_{PM_{2.5}}$) and 65.80\%($AQI_{NO_{2}}$) in India after the lockdown.

Other related works, include, the \cite{Kavouras2021_COVID-19}, in which the densities of $CO$, $O_3$, $NO_2$ and $SO_2$ were
predicted one week ahead, while, also, an AQI was calculated, which estimated that in countries, such as India, where the lockdown
measures were stricter, the air quality improved, while on the other hand in countries like Australia where the lockdown measures,
weren't so strict the air quality followed an incremented trend. In Madrid and Barcelona (Spain) \cite{baldasano_covid_19_2020_SpainPollution} 
the concentration of $NO_{2}$ reduced 62\% and 50\%, respectively, after a 75\% in traffic volume. In Baghdad (Iraq) \cite{HASHIM2021141978} 
the daily densities of $NO_2$, $O_3$, $PM_{2.5}$ and $PM_{10}$ used before and during the lockdown, in addition to the AQI for the same time period. 

\section{Proposed methodology}
\label{Proposed_methodology}
In this work we focus on two research topics: a) how each of the specific prevention measures is related on each of the pollutant factor levels, and b) to what extent we could estimate (forecast) the trend of each pollutant level, when using as input the current level, plus the prevention measures. On the one hand, the estimation of the correlation between COVID-19 measures and pollutant factors can be evaluated using various case-specific techniques \cite{eftychios2018correlationMethods}. On the other hand, the estimation of pollutant factors' levels few days ahead, entails to traditional regression problem 
\cite{kavouras2018skeleton}. 

Let us denote as $\textbf{x}=[x_1, ..., x_n]$, a set of predictive factors, involving prevention measures magnitude and current pollution levels, and $\textbf{y}$ as the investigated pollutant factor. We will try to estimate a predictive function $f(\textbf{x}) \longrightarrow \hat{\textbf{y}}$, where $\hat{y}$ stands for the specific pollutant factor's level, two days ahead. Practically, we would like to have values as close as possible to actual ones, i.e. $\hat{\textbf{y}} \simeq \textbf{y}$ 
\cite{kopsiaftis2019gaussian}. 


The input data, is a combination of: a) the densities of the most harmful pollutant gases ($CO$, $O_3$, $NO_2$ and $SO_2$), day-by-day period, and 
b) the magnitude of the applied prevention measure ranging from 0 (no measure applied) to 1 (the strictest form of the measure). Experiments rung in 
a total of 4 European Cities (Athens, Gladsaxe, Lodz, Rome). The adopted COVID-19 mitigation strategies, considered as predictive factors, are explained 
in the following lines. 

\textbf{Restrictions of Internal Movement} (RE\_IN\_MOV) refers to the transportation between municipalities.
\textbf{International Travel Controls} (IN\_TR\_CON) include any measure which restricts/prohibits the transportation between countries.
\textbf{Cancellation of Public Events} (CA\_PUB\_EV) involves the cancellation of all public events, locally (e.g. football matches, concerts) 
or worldwide (e.g. Olympic Games in Japan, Euro-vision).
\textbf{Restriction in Gatherings} (RE\_GAT) is a general description for all measures that restricts more than a number of people be together, in the same place.
\textbf{Close Public Transportation} (C\_PUB\_TRAN) includes the measures taken about the closure of public transportation (e.g. busses, metro).
\textbf{School Closures} (C\_SCHOOL) describe any measure involving the education system (e.g. school closure, distance learning, etc).
\textbf{Stay at Home Requirement}(STAY\_HOME\_R) is considered the most intense measure, since it penalized any movement outside home, without a specified reason 
(e.g. visit to the doctor, pharmacy, provide assistance to relative, etc).
\textbf{Workplace Closures} (C\_WORKPLACE) refer to both the industrial activity and the commercial sector. 

The COVID-19 dataset has been downloaded from Our World in Data (\url{https://ourworldindata.org/policy-responses-covid}) 
\cite{SITE_FOR_COVID_MEASURES_INTERNATIONALLY}, which been updated daily. The densities of the four pollutant factors have been downloaded 
from the Sentinel-5/TROPOMI dataset of Google Earth Engine \cite{SITE_FOR_S05_DATA}, in a two days period, with 50m spatial analysis.

\subsection{Investigating patterns similarity}
As a first step, two statistical methods have been used. The core idea was to identify whether there is high correlation between the investigated pollutant factor and any of the applied prevention measures.These statistical measures are known as a) Pearson correlation and b) Dynamic Time Wrapping (DTW).  

Pearson correlation estimates how related two signals are, using the best fit line approach. This     method calculates three values, which refers to correlation (r), coefficient of determination ($R^2$) and hypothesis (p).
    The r can be any value between -1 and 1 (the sign shows the direction) and absolute value can be translated as:
    \begin{itemize}
        \item 0.00-0.09 = No Correlation
        \item 0.10-0.39 = Weak Correlation
        \item 0.40-0.69 = Moderate Correlation
        \item 0.70-0.89 = Strong Correlation
        \item 0.90-1.00 = Very Strong Correlation
    \end{itemize}
    
    The $R^2$ can be any value between [0, 1] and can be multiplied by 100.0 to express the percent of affection between correlated objects. The p can be any value between range
    [0, 1] representing the probability that this data would have arisen if the null hypothesis were true (\textbf{Null hypothesis}: There is no correlation between A and B in the overall population $r=0$, \textbf{Alternative hypothesis}: There is a correlation between A and B in the overall population $r\neq0$). 
    \cite{PearsonMethod_1, PearsonMethod_2}
    
DTW is a technique used for calculating the alignment between two given (time-dependent) signals, under certain restrictions. The sequences are warped in a non-linear fashion to match each other. DTW, originally, has been used for the comparison of different speech patterns in automatic speech recognition. \cite{DynamicTimeWrapping}

\subsection{Machine Learning Tools for Extracting Environmental Impact } 
Various machine approaches have been evaluated. These approaches range from traditional shallow learning techniques like k-nearest neighbors to 
more complex approaches as deep neural networks. The outcomes indicate that, for the problem at hand, and due to the low feature space dimensional, traditional approaches are sufficient for robust pollutant factors level estimations.

\textbf{Deep Neural Network (DNN)}: A DNN method is a feed forward network that consists of multiple hidden layers. For the purpose of this research we used for input a set of predictive values $\textbf{x}=[x_1, ..., x_n]$, and as output $\textbf{y}=[y_1, y_2]$. The architecture, also, consists of three hidden layers ($L_1$, $L_2$, $L_3$), with sizes 20, 10 and 20 accordingly. As activation function between the hidden layers, the SELU (Scaled Exponential Linear Units) has been chosen, and for the output layer the sigmoid.

\textbf{Decision Tree Regressor (DTR)}: DTR method's core idea lies in sub-dividing the space into smaller regions
and then fit simple models to them.  Practically, every internal node tests an attribute and according to which cell the path, on the branches, ends, an average value over the available observations is calculated \cite{DecTree}. The main parameter a DTR need for running appropriately is the maximum depth of the tree. In our case we used a $maximumDepth = 5$.

\textbf{Random Forest Regression(RFR)} RFR is an alternative of the DTR (in case of regression can be called regression tree -  RTR - as well) methodology. In this case the concept of the DTR(RTR) can be extended using the power of contemporary computers to generate hundreds of DTRs(RTRs), simultaneously, known as random forests.  The final prediction can be estimated by the smallest MSE calculated from the average of the various DTRs(RTRs)\cite{SMITH201385, wang_four_month_2020_RFI}. RFR takes as parameters the maximum depth
and a random state. We used $maximumDepth = 5$ and $randomState = 2$.

\textbf{K-Nearest Neighbor (K-NN)}: K-NN  is a non parametric supervised machine learning method, which makes no assumptions for the underlying data distribution. In this method, for every instance, the distances (usually Euclidean) between a $x_i$ feature and all features of the training set,  are calculated. Then the k-nearest 
neighbors are selected and the $x_i$ feature is classified with the most frequent class among the k-nearest neighbors \cite{KNN}.
    
\textbf{Linear Regression (LReg)} can be expressed as the statistical method applied to a dataset to quantify and define the relation between the considered features. LReg can be used for forecasting. We have, also, consider other types of linear models, fitted by minimizing a regularized empirical loss with \textbf{Stochastic Gradient Descent (Multi O/P GBR - MGBR)}, or by fitting a regressor on the original dataset and then fits additional copies of the regressor on the same dataset but where the weights of instances are adjusted according to the error of the current prediction. As such, subsequent regressors focus more on difficult cases  (\textbf{Multi O/P AdaB - MAdaB}) \cite{MISHRA2020949}. In both training methods (MGBR and MAdab) we used $estimators=5$. 

\textbf{Lasso} is a modification of linear regression, where the model is penalized for the sum of absolute values of the weights. Thus, the absolute values of weight will be (in general) reduced, and many will tend to be zeros. Lasso algorithm can be explained further by the LARS algorithm \cite{Lasso}. \textbf{Ridge} regression is another method for modeling the connection between a depended scalar variable with one or more explanatory variables. It can be seen as a step further to Lasso. Ridge regression penalizes the model for the sum of squared value of the weights. Thus, the weights not only tend to have smaller absolute values, but also really tend to penalize the extremes of the weights, resulting in a group of weights that are more evenly distributed \cite{RIDGE}.

\section{Experimental Setup}
\subsection{Dataset description}
\label{experimental_setup}
The dataset is composed by 4 European Cities (Athens, Gladsaxe, Lodz, Rome). 
For each city has been calculated the measures per two days period from the 
1st of January, 2020, as the first day and 31st of December, 2020, as the last.
This methodology splits the year into 183 periods (leap year) of two days. For 
each period has been calculated the mean values of the COVID-19 measures 
described in Section \ref{Proposed_methodology}. The raw COVID-19 measures 
dataset can take integer values between 0 (no measures) to 4 (strict 
restriction). These values have been normalized in range [0-no measure, 1-strict measure], over a two day period calculation (average).

The pollutant concentrations of the most harmful gases ($CO$, $O_3$, $NO_2$ and $SO_2$) has been downloaded from Sentinel-5P/TROPOMI satellite from Google 
Earth Engine API \cite{SITE_FOR_S05_DATA}. To be more specific only the density band, which represent the density of pollutant per pixel in $mol/m^2$ 
unit has been downloaded. As clipped geometry has been chosen a square of $0.50deg$ with center coordinates the longitude and latitude of the city center, with 
a pixel size of $50m\times50m$. The final pixel density has been calculated as the mean density per period. Furthermore for the linear regression to run the 
image needed to be transformed as a scalar value. The transformation achieved by calculated the mean and standard deviation per city for each of the four pollutants.

Finally, the two datasets combined and for each country the train/test input compromised from the 8 COVID-19 measures and one pollutant, out of four in total, (mean and 
standard deviation for the current week), while the train/test output compromised from the mean and standard deviation for the next week. In the end, experiments were based
on a dataset of size $10$ feature values $\times$ 182 periods $\times$ 4 cities. The same dataset has been, also, used to run the correlation tests.

\subsection{Experimental results}

\begin{figure*}[!ht]
    \centering
    \includegraphics[width=1.0\linewidth]{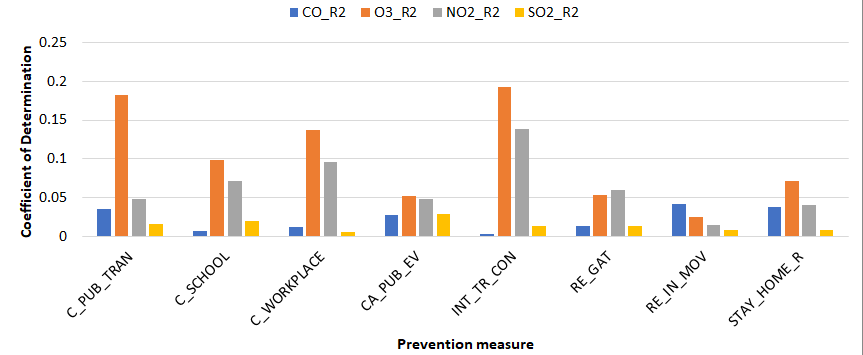}
    \caption{Coefficient of Determination ($R^2$) from Pearson Correlation}
    \label{fig_R2}
\end{figure*}

Figure \ref{fig_R2} indicates the coefficient of determination (CoD). The first observation is that all measures have $CoD < 20\% (0.20)$, which indicates weak to moderate correlation for most cases. In particular, internal travel restrictions had a moderate correlation to NO2 levels. Yet, the same restrictions appear to have no correlation to $CO$ levels. Restrictions of Internal Movement seems to have no correlation with any of the pollutant factors' levels.

\begin{figure*}[!ht]
    \centering
    \includegraphics[width=1.0\linewidth]{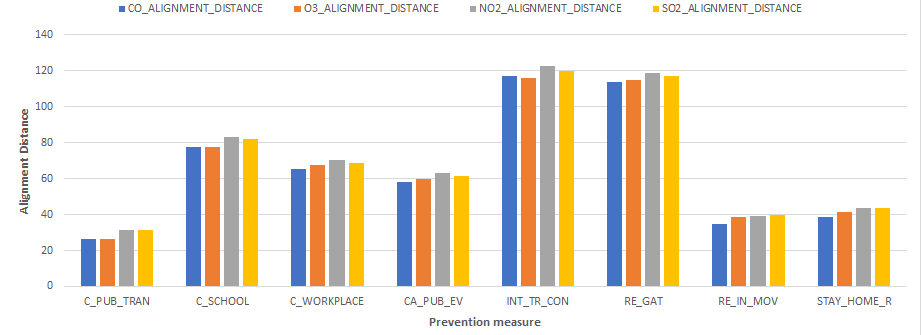}
    \caption{Alignment Distance from Dynamic Time Warping}
    \label{fig_DTW}
\end{figure*}

Figure \ref{fig_DTW} presents the alignment distance between the investigated pollutant factor and the applied measure. 
A small distance suggest that the behavioural patterns are more related. As such, closing the public transportation or 
imposing restrictions in movements signal patterns are closer to the ones of the pollutant factors. In this case the alignment distance was at least two times lesser compared to the one of international travel controls. The same is observed for the restriction in gatherings.

\begin{figure*}[!ht]
    \centering
    \includegraphics[width=1.0\linewidth]{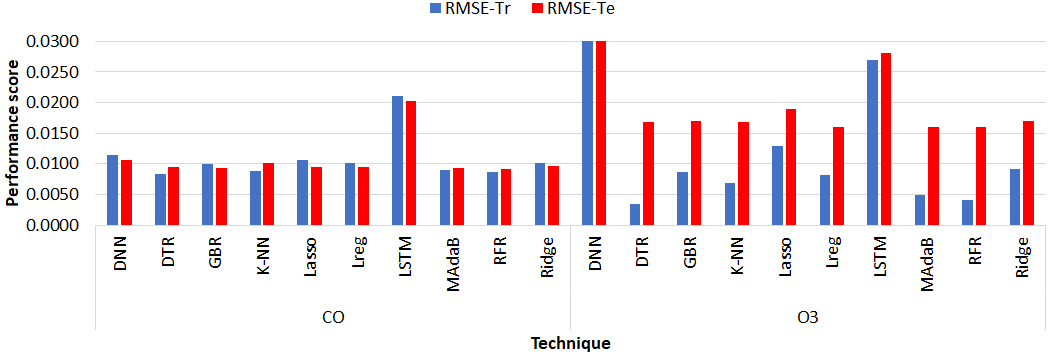}
    \caption{Root Mean Square Error (RMSE) for each machine and deep learning techniques for $CO$ and $O_3$.}
    \label{fig_RMSE_CO_O3}
\end{figure*}

\begin{figure*}[!ht]
    \centering
    \includegraphics[width=1.0\linewidth]{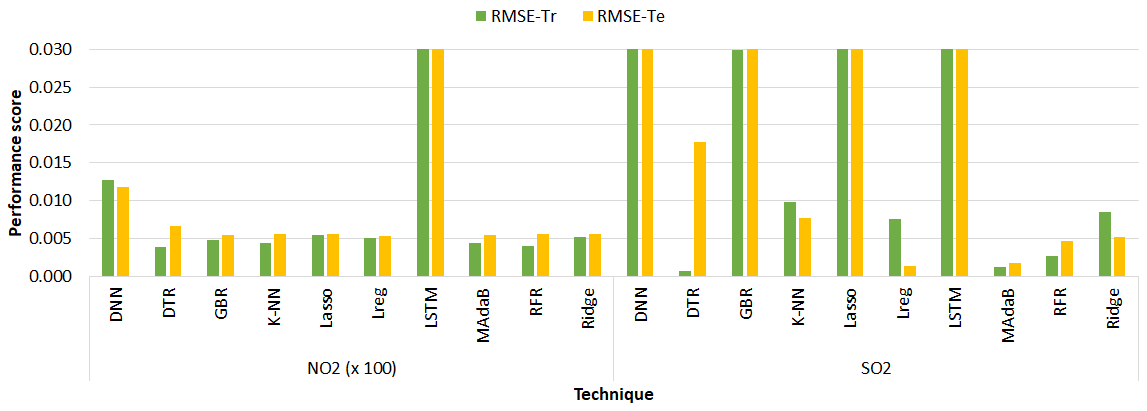}
    \caption{Root Mean Square Error (RMSE) for each machine and deep learning techniques for $NO_2$ and $SO_2$.}
    \label{fig_RMSE_NO2_SO2}
\end{figure*}

Taking into account the previous analysis, we decided to use all measures for the training/testing the machine/deep learning techniques. From Figure \ref{fig_RMSE_CO_O3}
we can safely assume that RFR performs slightly better than the other techniques for both Carbon Monoxide($CO$) and Ozone($O_3$). Especially the DNN has the worst results for predicting the Ozone. In case of Nitrogen Dioxide and Sulphur Dioxide prediction (Figure \ref{fig_RMSE_NO2_SO2}), the best results observed in MAdaB technique. In this point we need to specify that the errors maybe seems low, however the difference with the input values (pollutant gases concentration) is in a 10\% scale (e.g. if the original value was 0.000030 the output value could be 0.000028, in case of $NO_2$).

\section{Conclusions}
In this work we indicated that a weak correlation between the COVID-19 and the behaviour of the four most dangerous pollutant gases exists. To prove this correlation
we performed a series of tests using Pearson and Dynamic Time Warping techniques for each of the 4 European Cities, using the each two days period COVID-19 measure 
values and each pollutant gas density. Figures \ref{fig_R2} and \ref{fig_DTW}, which depict the coefficient of determination and the alignment distance, respectively, show
that there is no measure with strong correlation, thus we used all the measures for the machine and deed learning models.

The machine learning outputs were sufficient enough for future predictions of the densities for each pollutant gas in a two days period. However, using bigger 
dataset the results can be further improved. Models like these, can be used, in the near future, for the estimation of the benefits from the replacement of 
petrol and oil vehicles, with other environmental friendly vehicles.


\section*{Acknowledgement}
This paper is supported by the European Union Funded project euPOLIS "Integrated NBS-based Urban Planning Methodology for Enhancing the Health and Well-being of Citizens: 
the euPOLIS Approach" under the Horizon 2020 program H2020-EU.3.5.2., grant agreement No 869448.

\end{document}